\documentclass[11pt,a4paper]{article}
\usepackage{times}
\usepackage{latexsym}
\usepackage{tikz}
\usepackage[ruled,vlined]{algorithm2e}
\usepackage{booktabs}
\usepackage{amsmath}
\usepackage{graphicx}
\usepackage{pgfplots}
\usepgfplotslibrary{colorbrewer}
\usetikzlibrary{plotmarks}
\PassOptionsToPackage{breaklinks}{hyperref}
\usepackage{hyperref}
\usepackage[hyperref]{acl2020}
\usepackage[hyphenbreaks]{breakurl}

\aclfinalcopy 

\usepackage{xargs}                      

\usepackage{todonotes}
\newcommandx{\unsure}[2][1=]{\todo[linecolor=red,backgroundcolor=red!25,bordercolor=red,#1]{#2}}
\newcommandx{\change}[2][1=]{\todo[linecolor=blue,backgroundcolor=blue!25,bordercolor=blue,#1]{#2}}
\newcommandx{\info}[2][1=]{\todo[linecolor=OliveGreen,backgroundcolor=OliveGreen!25,bordercolor=OliveGreen,#1]{#2}}
\newcommandx{\improvement}[2][1=]{\todo[linecolor=Plum,backgroundcolor=Plum!25,bordercolor=Plum,#1]{#2}}
\newcommandx{\thiswillnotshow}[2][1=]{\todo[disable,#1]{#2}}

\title{Bertrand-DR: Improving Text-to-SQL using a Discriminative Re-ranker}

\author{Amol Kelkar, Rohan Relan, Vaishali Bhardwaj\\ \textbf{Saurabh Vaichal}, \textbf{Chandra Khatri}, \textbf{Peter Relan} \\ Got It AI R\&D \\ {\tt \{amol,rohan,vaishali,saurabh,chandra,peter\}@got-it.ai}}

\date{}

\begin{document}
\maketitle

\begin{abstract}
To access data stored in relational databases, users need to understand the database schema and write a query using a query language such as SQL. To simplify this task, text-to-SQL models attempt to translate a user's natural language question to corresponding SQL query. Recently, several generative text-to-SQL models have been developed. We propose a novel discriminative re-ranker to improve the performance of generative text-to-SQL models by extracting the best SQL query from the beam output predicted by the text-to-SQL generator, resulting in improved performance in the cases where the best query was in the candidate list, but not at the top of the list. We build the re-ranker as a schema agnostic BERT fine-tuned classifier. We analyze relative strengths of the text-to-SQL and re-ranker models across different query hardness levels, and suggest how to combine the two models for optimal performance. We demonstrate the effectiveness of the re-ranker by applying it to two state-of-the-art text-to-SQL models, and achieve top 4 score on the Spider leaderboard at the time of writing this article.
\end{abstract}

\section{Introduction}
Data is generally stored in structured form through relational databases \cite{RelationalDatabase:1970}. SQL is the domain specific language used to query and manage data stored in most commercially available relational databases \cite{SQLPopular:1998}. Although powerful, SQL is hard to master and thus out of reach for many users who need to query data. As a result, various Natural Language Interfaces to Databases (NLIs) \cite{NLI:1987} have been investigated. One approach towards NLI is to use a text-to-SQL semantic parser to convert a natural language utterance into a SQL query. With the recent developments related to deep learning, a large number of neural network based models such as IRNet \cite{IRNet:Guo2019}, and GNN \cite{GNN:Bogin2019} have been developed for this task.

Many recent text-to-SQL models use a sequence-to-sequence approach, encoding a sequence of natural language tokens and then decoding a sequence of SQL tokens \cite{Grammar:2019}. During each decoding step, these models produce a probability distribution over a set of tokens such as SQL query tokens, production rules \cite{GNN:Bogin2019} or IR tokens \cite{IRNet:Guo2019}. The output SQL is produced either using a greedy best-first search \cite{GreedySearch:1960}, i.e. producing the most probable token at each time step, or using a beam search \cite{BeamSearch:1977}, i.e. keeping a set of most probable sequences of tokens and selecting the most probable sequence at the end.

A text-to-SQL parser is a generative model of the joint probability distribution \cite{generative_discriminative:2002} between natural language questions, schema information and correct SQL queries. This is a very complex task and models have had limited success unless the problem space is constrained using various techniques such as using intermediate representation \cite{IRNet:Guo2019} or using a grammar-based decoder \cite{Grammar:2019}. Here we explore a novel approach to improve the performance of text-to-SQL models.

Intuitively, it is significantly easier to recognize correct SQL from a set of candidate queries compared to writing correct SQL for a given utterance and schema. So, rather than putting all the burden on the generative model and the greedy/beam search for producing the correct query,  we can use a discriminator to perform a better search among candidate queries produced by the generator. Also, compared to a generative model, discriminators are relatively easier to train and have better accuracy because they need to model only the conditional probability distribution instead of the joint probability distribution \cite{generative_discriminative:2002}.

When used with a beam search \cite{BeamSearch:1977}, a text-to-SQL model produces a set of candidate SQL queries. It is observed that the correct query may be in the set of candidates, but not at the top of the list when sorted by token generation probabilities. We use a discriminative re-ranker \cite{Discriminative:2005} to identify the correct query from the set of candidates. It significantly improves the performance of text-to-SQL models. For example, our re-ranker improves SQL logical form accuracy of GNN \cite{GNN:Bogin2019} from 51.3\% to 57.9\%, when trained on the Spider \cite{Spider:Yu2018} dataset.

Some of the main contributions of this paper include: 1) We present a novel discriminative re-ranker for the text-to-SQL task, 2) We show that the re-ranker can improve multiple state-of-the-art text-to-SQL models, and 3) We demonstrate how to pair a text-to-SQL model with corresponding re-ranker for best performance.

We make the code for our model available at {\small \url{https://github.com/amolk/Bertrand-DR}}.

\section{Problem definition and notation}
Given a natural language utterance $u$ and schema information $s$, a text-to-SQL model generates a valid SQL query $q$ such that when executed against a database conforming to schema $s$, the query $q$ would produce answer to the question $u$.

text-to-SQL semantic parsing can be framed as a supervised learning task, to induce a function $\mathcal{F} : (\mathcal{U}, \mathcal{S}) \to \mathcal{Q}$ given each training example $(u, s, q)$, where $u \in \mathcal{U}$, the set of potential natural language utterances, $s \in \mathcal{S}$, the set of potential database schemas, and $q \in \mathcal{Q}$, the set of potential SQL queries. We define $\mathcal{G}(u, s) \subset \mathcal{Q}$ to be the set of candidates SQL queries generated for a given input $(u,s)$. The model assigns probability $\mathcal{P}(u, s, q)$ for each $(u,s)$ and $q$ pair. The most likely parse for each $(u,s)$ is
\begin{equation}
 	q_{greedy} = \operatorname*{argmax}_{q \in \mathcal{G}(u,s)} \mathcal{P}(u, s, q)
\end{equation}

$\mathcal{P}(u, s, q)$ is defined as the cumulative probability of the sequence of tokens $\{q_0, q_1, \ldots, q_t\}$ generated for $q$.

\begin{equation}
  	\mathcal{P}(u, s, q) = \prod_{i=0 \ldots t} \mathcal{P}(q_i | u, s, \{q_0, \ldots, q_{i-1}\})
\end{equation}

It is not practical to search through all possible candidate SQL queries $\mathcal{G}(u, s)$, so a beam search is often used to produce a subset of approximately best candidates $\mathcal{C} \subset \mathcal{G}(u, s)$. The candidate with the highest probability $q_{top} \in \mathcal{C}$ is used as the prediction of the model. The correct query $q_{gold}$ may be 

\begin{itemize}
\item at the top of $\mathcal{C}$, i.e. $q_{gold} = q_{top}$,
\item somewhere else within $\mathcal{C}$, i.e. $q_{gold} \in \mathcal{C}$ and $q_{gold} \neq q_{top}$, OR
\item it may not be present in the candidates list, i.e. $q_{gold} \notin \mathcal{C}$.
\end{itemize}

The first two cases together are considered a "beam hit" and the last case is a "beam miss". First case represents correct prediction by the text-to-SQL model. A discriminative re-ranker can be trained to improve accuracy in the second case, where $q_{gold}$ was predicted but it isn't at the top of the beam.

\section{Our approach}
We build Bertrand-DR, a Discriminative Re-ranker as a binary classifier by fine tuning BERT \cite{BERT:Devlin2019}, to predict whether a given candidate query $q$ is the gold query for given utterance $u$ and schema information $s$. 

First, utterance $u$ and query $q$ are encoded using BERT. We use {\tt [SEP]} to separate utterance and the query. We use WordPiece \cite{WordPiece2016} to tokenize utterance into utterance tokens $u_1...u_N$ and query into query tokens $q_1...q_M$, where $N$ and $M$ are the token counts for utterance and query, respectively. The tokens are combined as follows to form the input token sequence:

\medskip
{\tt [CLS], $u_1$, $u_2$, ... $u_N$, [SEP], $q_1$, $q_2$, ... $q_M$, [SEP]}

\medskip
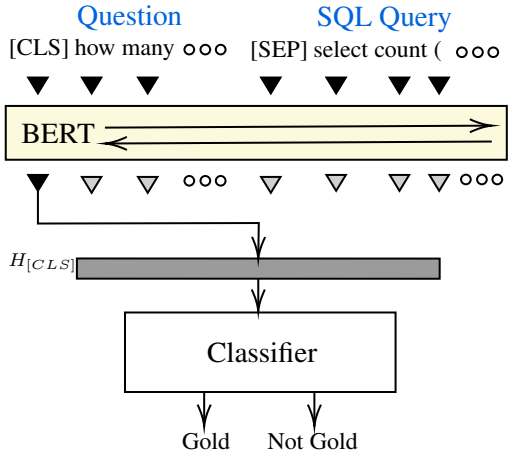
\begin{figure}[t]
\centering
\tikzset{every picture/.style={line width=0.75pt}} 

\begin{tikzpicture}[x=0.75pt,y=0.75pt,yscale=-1,xscale=1]

\draw  [fill={rgb, 255:red, 252; green, 249; blue, 220 }  ,fill opacity=1 ] (160,76.09) -- (410,76.09) -- (410,103.16) -- (160,103.16) -- cycle ;
\draw    (209.5,87) -- (400.5,86.01) ;
\draw [shift={(402.5,86)}, rotate = 539.7] [color={rgb, 255:red, 0; green, 0; blue, 0 }  ][line width=0.75]    (10.93,-3.29) .. controls (6.95,-1.4) and (3.31,-0.3) .. (0,0) .. controls (3.31,0.3) and (6.95,1.4) .. (10.93,3.29)   ;

\draw    (211.5,94.99) -- (402.5,94) ;

\draw [shift={(209.5,95)}, rotate = 359.7] [color={rgb, 255:red, 0; green, 0; blue, 0 }  ][line width=0.75]    (10.93,-3.29) .. controls (6.95,-1.4) and (3.31,-0.3) .. (0,0) .. controls (3.31,0.3) and (6.95,1.4) .. (10.93,3.29)   ;
\draw   (249.17,47.75) .. controls (249.17,46.42) and (250.25,45.33) .. (251.58,45.33) .. controls (252.92,45.33) and (254,46.42) .. (254,47.75) .. controls (254,49.08) and (252.92,50.17) .. (251.58,50.17) .. controls (250.25,50.17) and (249.17,49.08) .. (249.17,47.75) -- cycle ;
\draw   (257.17,47.75) .. controls (257.17,46.42) and (258.25,45.33) .. (259.58,45.33) .. controls (260.92,45.33) and (262,46.42) .. (262,47.75) .. controls (262,49.08) and (260.92,50.17) .. (259.58,50.17) .. controls (258.25,50.17) and (257.17,49.08) .. (257.17,47.75) -- cycle ;
\draw   (264.83,47.75) .. controls (264.83,46.42) and (265.92,45.33) .. (267.25,45.33) .. controls (268.58,45.33) and (269.67,46.42) .. (269.67,47.75) .. controls (269.67,49.08) and (268.58,50.17) .. (267.25,50.17) .. controls (265.92,50.17) and (264.83,49.08) .. (264.83,47.75) -- cycle ;

\draw   (249.17,113.75) .. controls (249.17,112.42) and (250.25,111.33) .. (251.58,111.33) .. controls (252.92,111.33) and (254,112.42) .. (254,113.75) .. controls (254,115.08) and (252.92,116.17) .. (251.58,116.17) .. controls (250.25,116.17) and (249.17,115.08) .. (249.17,113.75) -- cycle ;
\draw   (257.17,113.75) .. controls (257.17,112.42) and (258.25,111.33) .. (259.58,111.33) .. controls (260.92,111.33) and (262,112.42) .. (262,113.75) .. controls (262,115.08) and (260.92,116.17) .. (259.58,116.17) .. controls (258.25,116.17) and (257.17,115.08) .. (257.17,113.75) -- cycle ;
\draw   (264.83,113.75) .. controls (264.83,112.42) and (265.92,111.33) .. (267.25,111.33) .. controls (268.58,111.33) and (269.67,112.42) .. (269.67,113.75) .. controls (269.67,115.08) and (268.58,116.17) .. (267.25,116.17) .. controls (265.92,116.17) and (264.83,115.08) .. (264.83,113.75) -- cycle ;

\draw  [fill={rgb, 255:red, 0; green, 0; blue, 0 }  ,fill opacity=1 ] (176.05,69.8) -- (171.2,61.8) -- (180.9,61.8) -- cycle ;
\draw  [fill={rgb, 255:red, 0; green, 0; blue, 0 }  ,fill opacity=1 ] (202.85,69.8) -- (198,61.8) -- (207.7,61.8) -- cycle ;
\draw  [fill={rgb, 255:red, 0; green, 0; blue, 0 }  ,fill opacity=1 ] (230.85,69.8) -- (226,61.8) -- (235.7,61.8) -- cycle ;
\draw  [fill={rgb, 255:red, 0; green, 0; blue, 0 }  ,fill opacity=1 ] (292.25,69.8) -- (287.4,61.8) -- (297.1,61.8) -- cycle ;
\draw  [fill={rgb, 255:red, 0; green, 0; blue, 0 }  ,fill opacity=1 ] (324.85,69.8) -- (320,61.8) -- (329.7,61.8) -- cycle ;
\draw  [fill={rgb, 255:red, 0; green, 0; blue, 0 }  ,fill opacity=1 ] (356.45,69.8) -- (351.6,61.8) -- (361.3,61.8) -- cycle ;
\draw  [fill={rgb, 255:red, 0; green, 0; blue, 0 }  ,fill opacity=1 ] (376.25,69.8) -- (371.4,61.8) -- (381.1,61.8) -- cycle ;
\draw  [fill={rgb, 255:red, 0; green, 0; blue, 0 }  ,fill opacity=0.19 ] (292.25,119.8) -- (287.4,111.8) -- (297.1,111.8) -- cycle ;
\draw  [fill={rgb, 255:red, 0; green, 0; blue, 0 }  ,fill opacity=0.19 ] (324.85,118.8) -- (320,110.8) -- (329.7,110.8) -- cycle ;
\draw  [fill={rgb, 255:red, 0; green, 0; blue, 0 }  ,fill opacity=0.19 ] (356.45,118.8) -- (351.6,110.8) -- (361.3,110.8) -- cycle ;
\draw  [fill={rgb, 255:red, 0; green, 0; blue, 0 }  ,fill opacity=0.19 ] (376.25,118.8) -- (371.4,110.8) -- (381.1,110.8) -- cycle ;
\draw   (385.17,49.08) .. controls (385.17,47.75) and (386.25,46.67) .. (387.58,46.67) .. controls (388.92,46.67) and (390,47.75) .. (390,49.08) .. controls (390,50.42) and (388.92,51.5) .. (387.58,51.5) .. controls (386.25,51.5) and (385.17,50.42) .. (385.17,49.08) -- cycle ;
\draw   (393.17,49.08) .. controls (393.17,47.75) and (394.25,46.67) .. (395.58,46.67) .. controls (396.92,46.67) and (398,47.75) .. (398,49.08) .. controls (398,50.42) and (396.92,51.5) .. (395.58,51.5) .. controls (394.25,51.5) and (393.17,50.42) .. (393.17,49.08) -- cycle ;
\draw   (400.83,49.08) .. controls (400.83,47.75) and (401.92,46.67) .. (403.25,46.67) .. controls (404.58,46.67) and (405.67,47.75) .. (405.67,49.08) .. controls (405.67,50.42) and (404.58,51.5) .. (403.25,51.5) .. controls (401.92,51.5) and (400.83,50.42) .. (400.83,49.08) -- cycle ;

\draw   (387.17,113.08) .. controls (387.17,111.75) and (388.25,110.67) .. (389.58,110.67) .. controls (390.92,110.67) and (392,111.75) .. (392,113.08) .. controls (392,114.42) and (390.92,115.5) .. (389.58,115.5) .. controls (388.25,115.5) and (387.17,114.42) .. (387.17,113.08) -- cycle ;
\draw   (395.17,113.08) .. controls (395.17,111.75) and (396.25,110.67) .. (397.58,110.67) .. controls (398.92,110.67) and (400,111.75) .. (400,113.08) .. controls (400,114.42) and (398.92,115.5) .. (397.58,115.5) .. controls (396.25,115.5) and (395.17,114.42) .. (395.17,113.08) -- cycle ;
\draw   (402.83,113.08) .. controls (402.83,111.75) and (403.92,110.67) .. (405.25,110.67) .. controls (406.58,110.67) and (407.67,111.75) .. (407.67,113.08) .. controls (407.67,114.42) and (406.58,115.5) .. (405.25,115.5) .. controls (403.92,115.5) and (402.83,114.42) .. (402.83,113.08) -- cycle ;

\draw  [fill={rgb, 255:red, 155; green, 155; blue, 155 }  ,fill opacity=1 ] (195.5,162.97) -- (195.5,152.82) -- (376.5,152.88) -- (376.5,163.03) -- cycle ;
\draw  [fill={rgb, 255:red, 255; green, 255; blue, 255 }  ,fill opacity=1 ] (219.5,180) -- (354.5,180) -- (354.5,220) -- (219.5,220) -- cycle ;
\draw    (286,162.93) -- (286,177.39) ;
\draw [shift={(286,179.39)}, rotate = 270] [color={rgb, 255:red, 0; green, 0; blue, 0 }  ][line width=0.75]    (10.93,-3.29) .. controls (6.95,-1.4) and (3.31,-0.3) .. (0,0) .. controls (3.31,0.3) and (6.95,1.4) .. (10.93,3.29)   ;

\draw  [fill={rgb, 255:red, 0; green, 0; blue, 0 }  ,fill opacity=1 ] (176.05,118.8) -- (171.2,110.8) -- (180.9,110.8) -- cycle ;
\draw  [fill={rgb, 255:red, 0; green, 0; blue, 0 }  ,fill opacity=0.19 ] (202.85,119.8) -- (198,111.8) -- (207.7,111.8) -- cycle ;
\draw  [fill={rgb, 255:red, 0; green, 0; blue, 0 }  ,fill opacity=0.19 ] (230.85,119.8) -- (226,111.8) -- (235.7,111.8) -- cycle ;
\draw    (176.05,118.8) -- (176.05,135.25) -- (286.25,135) -- (286.03,150.93) ;
\draw [shift={(286,152.93)}, rotate = 270.8] [color={rgb, 255:red, 0; green, 0; blue, 0 }  ][line width=0.75]    (10.93,-3.29) .. controls (6.95,-1.4) and (3.31,-0.3) .. (0,0) .. controls (3.31,0.3) and (6.95,1.4) .. (10.93,3.29)   ;

\draw    (259,219.93) -- (259,235.3) ;
\draw [shift={(259,237.3)}, rotate = 270] [color={rgb, 255:red, 0; green, 0; blue, 0 }  ][line width=0.75]    (10.93,-3.29) .. controls (6.95,-1.4) and (3.31,-0.3) .. (0,0) .. controls (3.31,0.3) and (6.95,1.4) .. (10.93,3.29)   ;

\draw    (314,219.93) -- (314,235.3) ;
\draw [shift={(314,237.3)}, rotate = 270] [color={rgb, 255:red, 0; green, 0; blue, 0 }  ][line width=0.75]    (10.93,-3.29) .. controls (6.95,-1.4) and (3.31,-0.3) .. (0,0) .. controls (3.31,0.3) and (6.95,1.4) .. (10.93,3.29)   ;

\draw (186,89.63) node   [align=left] {BERT};
\draw (203,47.63) node  [font=\small] [align=left] {{\footnotesize [CLS] how many}};
\draw (222,31.63) node  [color={rgb, 255:red, 74; green, 144; blue, 226 }  ,opacity=1 ] [align=left] {\textcolor[rgb]{0,0.39,0.87}{Question}};
\draw (288,200) node   [align=left] {Classifier};
\draw (260,244.71) node  [font=\footnotesize] [align=left] {Gold};
\draw (313.14,244.71) node  [font=\footnotesize] [align=left] {Not Gold};
\draw (178.5,155) node  [font=\scriptsize]  {$H_{[ CLS]}$};
\draw (332.75,48.63) node  [font=\small] [align=left] {{\footnotesize [SEP] select count ( \ }};
\draw (349,32.63) node  [color={rgb, 255:red, 0; green, 0; blue, 255 }  ,opacity=1 ] [align=left] {\textcolor[rgb]{0,0.39,0.85}{SQL Query}};

\end{tikzpicture}
\caption{Bertrand-DR network architecture}
\label{figure:architecture}
\end{figure}

{\tt [CLS]} and {\tt [SEP]} are special tokens for classification and context separation, as used in \cite{BERT:Devlin2019}. The token sequence is then encoded using BERT. We take the last layer's hidden state for the first token, $H_{[CLS]}$, as the input embedding. $H_{[CLS]}$ is passed through a linear layer with {\tt tanh} activation function to form a pooled encoding. The resulting encoding is passed through a linear classification layer and the model is trained using binary cross entropy loss.

During inference, given utterance $u$ and query $q$, the model produces $\mathcal{P}(q \vert u)$, the probability that the given query is correct for the utterance. Note that schema information is not used. The probability is used as the $score$ for the query.

The beam search candidates produced by a text-to-SQL model are sorted by the generation probability of each candidate. This is the initial ranking of the candidates. The goal of the re-ranker is to predict a high score value for $q_{gold}$, if it is present in the candidates list, and lower score values for non-gold queries, such that $q_{gold}$ would end up at the top of the list when re-ordered by descending score values. An ideal re-ranker would ensure that whenever $q_{gold}$ is present in the candidates list, it is re-ranked to the top of the list. Thus, beam hit rate represents the theoretical maximum possible performance of a re-ranker.

In some cases, the text-to-SQL model produces the gold query at the top of the beam, but the re-ranker erroneously promotes a different candidate to the top. Thus, fully relying on the re-ranker's scores results in lower accuracy. We use an empirically derived threshold value $t$, such that a candidate is moved higher in order only if it has significantly larger score, i.e. larger at least by threshold $t$ compared to the candidate immediately above it. See Algorithm \ref{alg:bertrand} and figure \ref{figure:threshold}.

\begin{algorithm}[t]
 \SetAlgoLined
 \caption{Re-ranking algorithm}
 \label{alg:bertrand}
 \KwData{List of candidate SQL queries $\mathcal{C}$, Utterance $u$, Schema information $s$, a re-ranker $model$, and threshold $t$}
 \KwResult{SQL queries list $\mathcal{C}$ with the best candidate at the top}
 \ForEach{$c \in \mathcal{C}$}{
  $c.score = {model}(u, c)$
 }

 \For{i = C.length to 1}{
  \If{$C_{i}.score \geq C_{i-1}.score + t$} {
    swap($C_i$, $C_{i-1}$)
  }
 }

\end{algorithm}

\section{Experiments and results}
\renewcommand{\thefootnote}{$\dagger$}

\begin{table}[b!]
\centering
\begin{tabular}{ l c r }
  \toprule
  System & Dev & Test \\
  \midrule
  GNN \cite{GNN:Bogin2019} & 40.7 & 39.4 \\
  \ \ \ \ + Re-impl \cite{GNNGlobal:Bogin2019} & 44.1 & -- \\
  \ \ \ \ + BERT (ours) & 51.3 & -- \\
  \ \ \ \ + Bertrand-DR (ours) & \textbf{57.9} & \textbf{54.6} \\
  \midrule
  EditSQL \cite{EditSQL:Zhang2019} & 36.4 & 32.9 \\
  \ \ \ \ + BERT \cite{EditSQL:Zhang2019} & 57.6 & 53.4 \\
  \ \ \ \ + Bertrand-DR (ours) & \textbf{58.5} & --  \\

  \bottomrule
\end{tabular}
\caption{Logical form accuracy of various models on Spider dev and test datasets. Note that results on the Spider test dataset are available only for Spider leaderboard submissions. }
\label{table:Results}
\end{table}

\renewcommand{\thefootnote}{\arabic{footnote}}
\begin{table*}[t!]
\centering
\begin{tabular}{ l l c c c c c }
  \toprule
  System & Threshold & Easy & Medium & Hard & Extra Hard & Overall \\
  \midrule
  GNN \cite{GNN:Bogin2019} & \\
  \ \ \ \ + BERT (ours) &  &70.4 & 54.1 & 35.6 & 28.2 & 50.7 \\
  \ \ \ \ + Bertrand-DR (ours) & 0.10 & 81.6 & 62.5 & 39.1 & 30.6 & 57.9 \\
  \midrule
  EditSQL & \\
  \ \ \ \ + BERT \cite{EditSQL:Zhang2019} &  & 78.0 & 60.0 & 45.4 & 33.5 & 57.5 \\
  \ \ \ \ + Bertrand-DR (ours) & 0.68 & 81.6 & 59.5 & 47.1 & 33.5 & 58.5 \\
  \bottomrule
\end{tabular}
\caption{Logical form accuracy at various query hardness levels on Spider dev dataset. The re-ranker improved GNN results at all hardness levels, but for EditSQL, only the easy queries show improvement. The threshold levels were determined empirically to achieve best Overall performance.}
\label{table:ResultsByHardnessLevels}
\end{table*}

We apply our re-ranker on beam candidates generated by two state-of-the-art Text-to-SQL models, GNN \cite{GNN:Bogin2019} and EditSQL \cite{EditSQL:Zhang2019}.

\subsection{GNN}
The GNN \cite{GNN:Bogin2019} model is a grammar based semantic parser that uses a Graph Neural Network \cite{GraphNeuralNetwork} to encode schema entity relationships. The GNN model uses learned word embeddings. First, we modify GNN to leverage BERT embeddings. We use the scheme proposed by \cite{Spider:Yu2018} to compute question and schema entity embeddings using BERT. These embeddings are then used by GNN to encode the question, schema entities and to compute the schema linking. This improves the SQL logical form accuracy of GNN on the Spider \cite{Spider:Yu2018} dev dataset from 44.1\% to 51.3\%.

We train GNN model on the Spider dataset and then collect all beam candidates it produces on training and dev sets, using beam size 40. The collected dataset of beam candidates serves as training and dev datasets for the re-ranker.

When the correct query is present in the set of beam candidates, it is counted as a beam hit. Beam hit rate is the percent examples that result in a beam hit and it represents the theoretical maximum possible performance by the re-ranker. Beam hit rate for GNN was 62.4\% and the re-ranker produced accuracy close to the theoretical max accuracy.

\begin{table}[b!]
\centering
\begin{tabular}{ l c c c }
  \toprule
   & Base & + Bertrand-DR & Beam hit \\
  \midrule
  GNN & 51.3 & 57.9 & 62.3 \\
  EditSQL & 57.6 & 58.5 & 78.7 \\

  \bottomrule
\end{tabular}
\caption{Base indicates base model logical form accuracy on Spider dev dataset. Beam hit is the theoretical maximum performance a re-ranker model could achieve.}
\label{table:BeamHit}
\end{table}

\subsection{EditSQL}
The EditSQL implementation by \cite{EditSQL:Zhang2019} uses greedy search strategy for the decoder. We replace that with a beam search. Our implementation is available at {\small \url{https://github.com/amolk/editsql/tree/beam_search}}. Using beam search improved the accuracy on the Spider dev dataset by a small amount, from 57.6\% to 57.7\%. Beam hit rate for EditSQL was 78.7\%.

Due to hardware constraints, instead of 40, we used beam size 10 on a trained EditSQL model to generate a training and dev set for the re-ranker. This sub-optimal dataset resulted in a re-ranker that did not make much progress towards achieving accuracy close to the beam hit rate.

\subsection{Re-ranker training}
The re-ranker uses the BERT-base model \cite{BERT:Devlin2019} with a hidden size of 768, 12 Transformer blocks \cite{Transformer:Vaswani2017} and 12 self-attention heads. We use a pre-trained BERT model from the PyTorch transformer repository \cite{Huggingface:Wolf2019}. We fine-tune the BERT model on a single V100 GPU with dropout probability 0.1. We use Adam optimizer \cite{Adam} with a learning rate 1e-3 for the linear classifier layer. A lower learning rate of 5e-6 is used to fine tune BERT layers. After training the re-ranker, we empirically find a good threshold value to maximize the accuracy indicated by the Spider \cite{Spider:Yu2018} evaluation scripts.

If the re-ranker is applied to other text-to-SQL models, such as those listed on the Spider leaderboard, we expect that each model would have different optimal threshold value.

\subsection{Threshold tuning}
\begin{figure}[h!]
\resizebox{1.0\linewidth}{!}{\begin{tikzpicture}
\begin{axis}[
    xlabel={Threshold},
    ylabel={\% improvement},
    xmin=0, xmax=1,
    ymin=-2, ymax=2,
    xmajorgrids=true,
    ymajorgrids=true,
    grid style=dashed,
    thick=true,
]
 
\addplot[
    color=black,
    mark=star,
    ]
    coordinates {
		(0.05, -1.22)
		(0.1, -1.22)
		(0.15, -1.57)
		(0.2, -0.35)
		(0.3, -0.70)
		(0.4, -0.17)
		(0.5, 0.52)
		(0.62, 1.57)
		(0.68, 1.74)
		(0.75, 1.57)
		(0.87, 1.22)
		(1, 0.35)
    };
 
\end{axis}
\end{tikzpicture}}
\caption{Re-ranker performance improvement for EditSQL at various threshold values. For EditSQL, threshold value 0.68 performs best.}
\label{figure:threshold}
\end{figure}
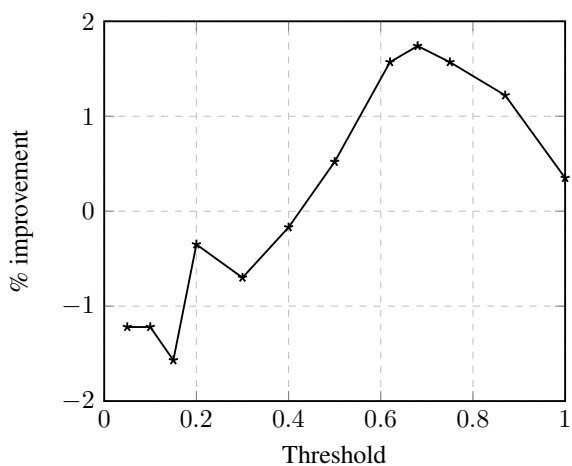

Re-ranker's performance depends on the threshold value. A threshold value of 0 allows re-ranker to promote a candidate if it scores more than the candidate above it. A threshold of $\infty$ turns off the re-ranker by disallowing all re-ranking decisions. A positive threshold value promotes a candidate only if its score is at least threshold larger than the score of the candidate above, thus allowing only high confidence re-ranking decisions to be executed. Figure \ref{figure:threshold} shows results of a threshold tuning experiment for EditSQL. Notice that the re-ranker performs best at the empirically determined threshold value 0.68.

We realize that this tuning process uses validation data which makes it vulnerable to overfitting, resulting in potentially poor performance on unseen test data. We mitigate this risk by using half of the validation data for tuning the threshold, while measuring performance on the other half.

\subsection{Results}

We compare results of BERT based text-to-SQL models with and without using the re-ranker. Our implementation of BERT enabled GNN has 51.3 accuracy on Spider dev set. Re-ranker improves that to 57.9, a 12.9\% relative improvement. The re-ranker, when trained using beam data from EditSQL, improves EditSQL's \cite{EditSQL:Zhang2019} 57.6 accuracy to 58.5, a 1.5\% relative improvement.

The results of our model can be seen on the Spider leaderboard {\small \url{https://yale-lily.github.io/spider}}. At the time of writing this artical, it was ranked 4th on the leaderboard.

\subsection{Analysis}
As seen in table \ref{table:ResultsByHardnessLevels}, the re-ranker improves GNN results at all hardness levels, but for EditSQL, only the {\small\tt easy} queries show improvement.

We trained the re-ranker using two different input configurations: 1) using utterance, schema and query, and 2) using only utterance and query. We observed that including schema information reduced performance at all hardness levels. We hypothesize that the model is better able to use the syntactic similarities between utterance and query and including schema information breaks that mapping, leading to worse performance.

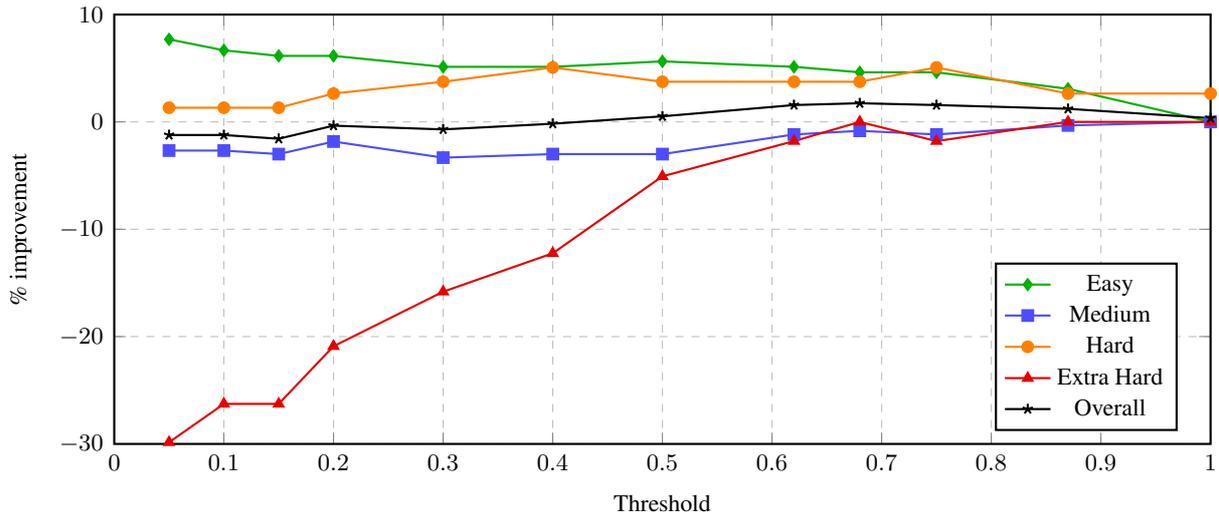
\begin{figure*}
\centering
\begin{tikzpicture}
\begin{axis}[
    xlabel={Threshold},
    ylabel={\% improvement},
    xmin=0, xmax=1,
    ymin=-30, ymax=10,
    xmajorgrids=true,
    ymajorgrids=true,
    grid style=dashed,
    thick=true,
    legend pos=south east,
    width=\textwidth,
    height=\axisdefaultheight,
    font=\small,
]

\addplot[
	color=green!70!black,
    mark=diamond*,
    ]
    coordinates {
	    (0.05, 7.69)
	    (0.1, 6.67)
	    (0.15, 6.15)
	    (0.2, 6.15)
	    (0.3, 5.13)
	    (0.4, 5.13)
	    (0.5, 5.64)
	    (0.62, 5.13)
	    (0.68, 4.62)
	    (0.75, 4.62)
	    (0.87, 3.08)
	    (1, 0.00)
    };
 
\addplot[
	color=blue!70!white,
    mark=square*
    ]
    coordinates {
		(0.05, -2.67)
		(0.1, -2.67)
		(0.15, -3.00)
		(0.2, -1.83)
		(0.3, -3.33)
		(0.4, -3.00)
		(0.5, -3.00)
		(0.62, -1.17)
		(0.68, -0.83)
		(0.75, -1.17)
		(0.87, -0.33)
		(1, 0.00)
    };

\addplot[
	color=orange,
    mark=oplus*
    ]
    coordinates {
		(0.05, 1.32)
		(0.1, 1.32)
		(0.15, 1.32)
		(0.2, 2.64)
		(0.3, 3.74)
		(0.4, 5.07)
		(0.5, 3.74)
		(0.62, 3.74)
		(0.68, 3.74)
		(0.75, 5.07)
		(0.87, 2.64)
		(1, 2.64)
    };

\addplot[
	color=red!90!black,
    mark=triangle*
    ]
    coordinates {
		(0.05, -29.85)
		(0.1, -26.27)
		(0.15, -26.27)
		(0.2, -20.90)
		(0.3, -15.82)
		(0.4, -12.24)
		(0.5, -5.07)
		(0.62, -1.79)
		(0.68, 0.00)
		(0.75, -1.79)
		(0.87, 0.00)
		(1, 0.00)
    };

\addplot[
    mark=star,
    ]
    coordinates {
		(0.05, -1.22)
		(0.1, -1.22)
		(0.15, -1.57)
		(0.2, -0.35)
		(0.3, -0.70)
		(0.4, -0.17)
		(0.5, 0.52)
		(0.62, 1.57)
		(0.68, 1.74)
		(0.75, 1.57)
		(0.87, 1.22)
		(1, 0.35)
    };

\legend{Easy\\Medium\\Hard\\Extra Hard\\Overall\\}

\end{axis}
\end{tikzpicture}

\caption{Re-ranker performance improvement for EditSQL at various threshold values for queries of various hardness levels.}
\label{figure:threshold_hardness}
\end{figure*}

What threshold value works best depends on the relative strengths of the text-to-SQL model and the corresponding re-ranker model. Spider evaluation scripts \cite{Spider:Yu2018} categorize SQL queries into 4 hardness buckets - {\small\tt easy}, {\small\tt medium}, {\small\tt hard} and {\small\tt extra hard} - based on the set of SQL features that appear in the queries. Figure \ref{figure:threshold_hardness} shows performance of the re-ranker for EditSQL for queries of various hardness levels at various threshold values.

At low threshold values, performance of {\small\tt easy} queries see most improvement, while performance of {\small\tt extra hard} queries suffers substantially. Looking at the queries, we noticed that {\small\tt easy} queries tend to look semantically similar to the corresponding utterance. In contrast, {\small\tt hard} and {\small\tt extra hard} queries have structures such as joins and sub-queries that depend on the database schema and do not resemble the utterance semantically.

We hypothesize that the re-ranker does better with {\small\tt easy} queries because it can use the semantic similarities between such utterance and query pairs. On the other hard, text-to-SQL models perform relatively better on harder queries where schema based information is utilized to select appropriate columns, joins and sub-queries. Also, text-to-SQL models trained on a complex dataset such as Spider are probably biased towards more complex questions and queries and thus perform worse on easy queries. The re-ranker may correct this bias.

The utterance words could be semantically related to the table and column names used in the corresponding queries. BERT is pre-trained on various language model tasks and is thus capable to mapping utterance words to table and column names if they are semantically related. In {\small\tt easy} queries, this mapping is more apparent. In contrast, {\small\tt hard} queries use constructs such as table joins that may not have direct semantic relationship with utterance words.

\section{Conclusion and Future work}
In this paper, we demonstrate a discriminative re-ranking based approach to improving text-to-SQL performance. We make the code available as a library at {\small \url{https://github.com/amolk/Bertrand-DR}}. We hope that the library would ease the adoption of discriminative re-ranking technique for text-to-SQL models and other encoder-decoder based generative models.

There are several potential avenues for further research. For instance, we built a single re-ranker model for queries of all hardness levels. Given various hardness level queries align differently in syntax and semantics with corresponding utterances, building separate re-ranker models for different hardness levels might allow the models to focus on relevant features. In such an ensemble, appropriate re-ranker could be selected based on the hardness level of the query at the top of the beam produced by the text-to-SQL model. Alternately, a separate model can be trained to predict query hardness for a given utterance and schema.

Another ensemble model approach would be to use other models that work well with {\small \tt easy} queries in conjunction with text-to-SQL models. WikiSQL dataset contains queries without JOINs and sub-queries, and thus tend to be {\small \tt easy} queries. Models that perform well on WikiSQL dataset, such as X-SQL \cite{x-sql:2019} and SQLova \cite{SQLova:2019} would be good candidates as the {\small \tt easy} query model.


We plan to apply our re-ranker to other state-of-the-art text-to-SQL models at the top of the Spider leaderboard. Current leading models such as IRNet \cite{IRNet:Guo2019} are well suited to be used with the re-ranker.

\section{Acknowledgements}
We thank Ben Bogin and Rui Zhang for helping us utilize GNN and EditSQL code bases, respectively. We also appreciate the efforts by Tao Yu in managing the Spider leaderboard and helping us with our leaderboard submission. 

Bertrand AI is a suite of AI models developed by Got It Inc to tackle the text-2-SQL problem. The model described in this paper, Bertrand-DR, is part of that suite. We thank the Got It team for providing valuable feedback in the process of productization of this work to build a conversational interface to databases.

\section{References}

\bibliographystyle{acl_natbib}
\bibliography{BERTRAND}


\end{document}